\documentclass[letterpaper, 10 pt, conference]{ieeeconf}  % Comment this line out if you need a4paper

\IEEEoverridecommandlockouts                              % This command is only needed if
\usepackage[pdftex]{graphicx}
\usepackage{multirow}
\usepackage{marvosym}
\usepackage{amsmath}
\usepackage{amssymb}
\usepackage{url}
\usepackage{cite}
\usepackage{makecell}
\usepackage{subfigure}
\usepackage[ruled,vlined]{algorithm2e}

\graphicspath{{Figure/}}

\title{\LARGE \bf
Rethinking Temporal Object Detection from Robotic Perspectives
}

\author{Xingyu Chen, Zhengxing Wu, Junzhi Yu, and Li Wen % <-this % stops a space
\thanks{This work was supported by the National Natural Science Foundation of China (nos. 61633004, 61633020, 61603388, 61633017, and 61725305), and by the Beijing Natural Science Foundation (no. 4161002).}% <-this % stops a space
\thanks{X. Chen, Z. Wu, and J. Yu are with the State Key Laboratory of Management and Control for Complex Systems, Institute of Automation, Chinese Academy of Sciences, Beijing 100190, China and University of Chinese Academy of Sciences, Beijing 100049, China. J. Yu is also with the Beijing Innovation Center for Engineering Science and Advanced Technology, Peking University, Beijing 100871, China
        {\tt\small \{chenxingyu2015, zhengxing.wu, junzhi.yu\}@ia.ac.cn}}
\thanks{L. Wen is with the School of Mechanical Engineering and Automation, Beihang University, Beijing 100191, China, {\tt\small liwen@buaa.edu.cn}}
        }

\begin{document}
\maketitle
\thispagestyle{empty}
\pagestyle{empty}

%%%%%%%%%%%%%%%%%%%%%%%%%%%%%%%%%%%%%%%%%%%%%%%%%%%%%%%%%%%%%%%%%%%%%%%%%%%%%%%%
\begin{abstract}
Video object detection (VID) has been vigorously studied for years but almost all literature adopts a static accuracy-based evaluation, i.e., average precision (AP). From a robotic perspective, the importance of recall continuity and localization stability is equal to that of accuracy, but the AP is insufficient to reflect detectors' performance across time. In this paper, non-reference assessments are proposed for continuity and stability based on object tracklets. These temporal evaluations can serve as supplements to static AP. Further, we develop an online tracklet refinement for improving detectors' temporal performance through short tracklet suppression, fragment filling, and temporal location fusion.

In addition, we propose a small-overlap suppression to extend VID methods to single object tracking (SOT) task so that a flexible SOT-by-detection framework is then formed.

Extensive experiments are conducted on ImageNet VID dataset and real-world robotic tasks, where the superiority of our proposed approaches are validated and verified. Codes will be publicly available.
\end{abstract}

%%%%%%%%%%%%%%%%%%%%%%%%%%%%%%%%%%%%%%%%%%%%%%%%%%%%%%%%%%%%%%%%%%%%%%%%%%%%%%%%
\section{Introduction}
\label{sec:intro}

{\noindent\bf Background.} As a temporal perception task, video object detection (VID) is one of the main research areas in computer vision and a fundamental tactic for robotic perception. Over recent years, we have witnessed the development of temporal object detection on ImageNet VID dataset \cite{bib:VID}. As the exclusive assessment, average precision (AP) has grown from less than $50\%$ \cite{bib:VideoTub} to over $80\%$ \cite{bib:DT}.

{\noindent\bf Problem \& motivation.}  Different from images, video frames are interrelated on time series. Thus, almost all VID methods leverage across-time information to improve detection performance \cite{bib:SeqNms,bib:TCNN,bib:DT,bib:DorT,bib:TRN,bib:TPN,bib:TSSD,bib:STSN,bib:HPVD}, but their metric AP only statically reflects image-based accuracy, failing in across-time evaluation. We deem that the AP would neglect some important characteristics in VID.
%Nevertheless, the AP is initially proposed , but it is still the exclusive assessment that evaluates the performance of temporal detection.

\begin{figure}[!t]
\begin{center}
\includegraphics[width=9cm]{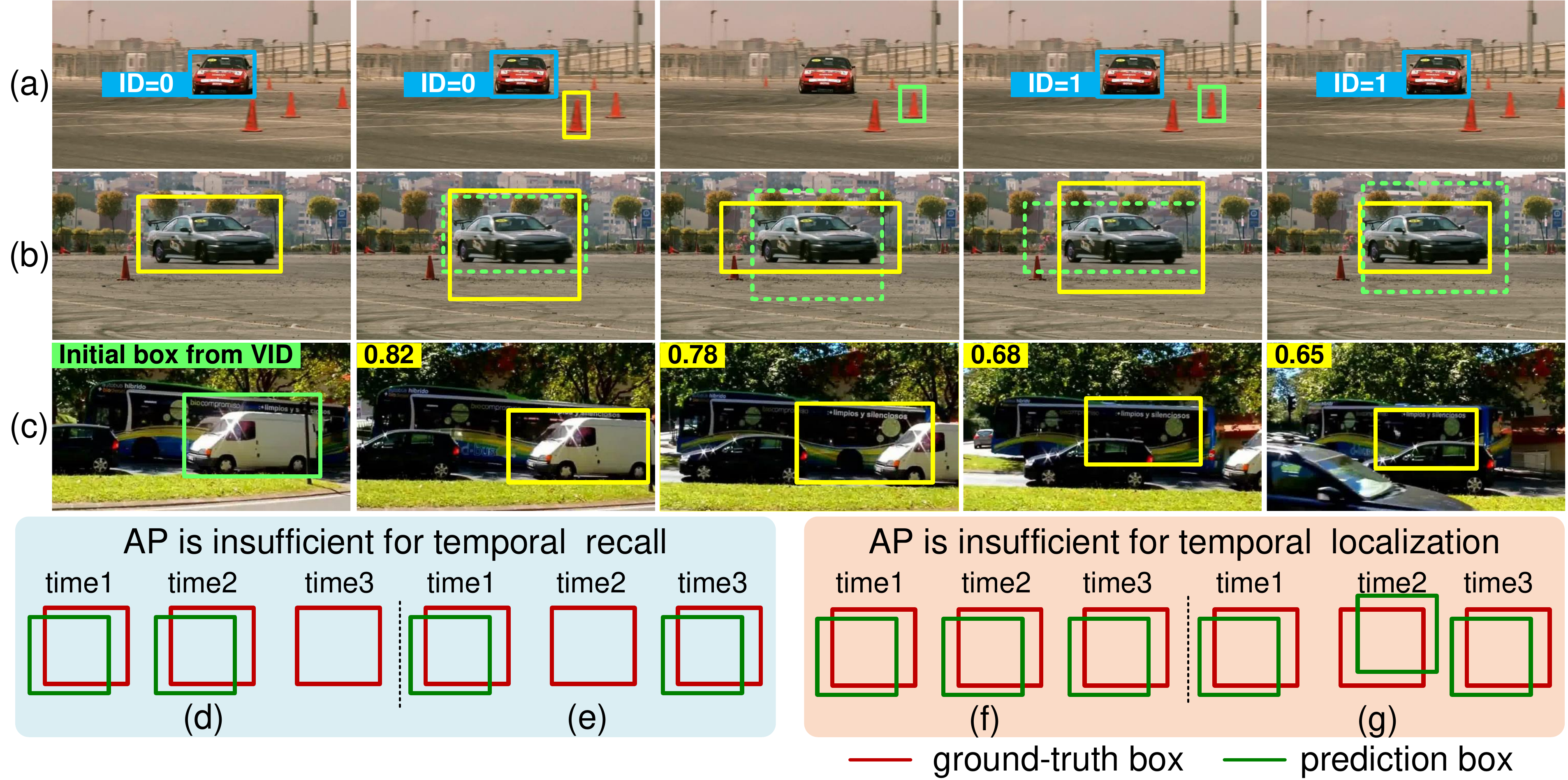}   % The printed column width is 8.4 cm.
\caption{Defective detection and tracking cases. (a) Short tracklet duration (yellow tracklet only contains 1 object while the length of green one is 2) and tracklet fragment; (b) box center/size jitter (green dashed boxes denote previous results); (c) siamese tracker needs an initial box from VID and is prone to drift (tracking score is shown in the left-top); (d--g) AP can hardly describe temporal recall/localization.}
\label{fig:intro}
\end{center}
\end{figure}

As shown in Fig.~\ref{fig:intro}~(a) and (b), besides false positive/negative captured by AP, defective cases of temporal detection are subdivided into two-fold aspects:
\begin{itemize}
  \item \emph{Recall continuity}: As shown in Fig.~\ref{fig:intro}~(a), transient object recall induces short tracklet duration, whose duration could only contain several frames. Additionally, intermittent object missing forms tracklet fragments, which could incur identity switch. We deem these phenomena damage recall continuity in VID.
  \item \emph{Localization stability}: As shown in Fig.~\ref{fig:intro}~(b), box center/size jitter frequently appears in modern object detection, and slight pixel-level change could incur considerable location jitter. It is conceived that this phenomenon impairs localization stability in VID.
\end{itemize}
Compared to detection accuracy, continuity and stability are equally important in robotic perception. Nevertheless, there has been relatively little work studying these two problems because the static evaluation system has difficulty in reflecting them. That is, AP performs not so well on describing detection continuity and stability. On one hand, although it reports object recall/missing from a spatial perspective, AP is insufficient for analysis of temporal classification. For example, AP can hardly distinguish cases in Fig.~\ref{fig:intro}~(d) and (e), but (d) is relatively better as for robotic perception. On the other hand, AP is also insufficient for localization jitter since intersection-over-union (IoU) cannot describe the direction of overlap. That is, AP is unable to discriminate cases in Fig.~\ref{fig:intro}~(f) and (g), but (f) is relatively better as for temporal localization.

Above problems inspire us to develop a temporal evaluation system for VID. We suggest a {\bf non-reference} manner for evaluation because 1) both continuity and stability are totally unrelated to man-made labels; 2) the evaluation system should be applied to various scenes (including but not limited to datasets), and most robotic scenarios lack ground-truth labels. As a relevant scope, multi-object tracking (MOT) aims to associate object boxes across time \cite{bib:MOT}, then object tracklets are produced. The tracklets from MOT exactly reflect VID performance across time, but there has been limited research on MOT-based VID analysis. This motivates us to re-investigate VID approaches, then generally evaluate and enhance detectors' continuity/stability with MOT.

VID-MOT deals with multiple tracklets in the manner of tracking-by-detection, but robotic perception also has an imperative need of single object tracking (SOT) \cite{bib:VOT}. If VID-MOT is directly leveraged for SOT, redundant computing would incur high time consumption. Additionally, researchers tend to exploit similarity-based SOT methods \cite{bib:KCF,bib:SiamRPN,bib:SiamFC++}. However, as shown in Fig.~\ref{fig:intro}~(c), similarity-based SOT is independent from VID and has three-fold drawbacks: 1) it is prone to suffer tracking drift; 2) It cannot work without initial box from VID \cite{bib:TDR}; 3) VID-SOT cascade would highly increase model complexity \cite{bib:CDT}. All these limitations are unfavorable for robotic perception, inspiring us to extend VID methods towards SOT task.

{\noindent\bf Our work.} In this paper, we firstly study temporal performance on object detection. We pose transient and intermittent object recall/missing as a problem of recall continuity, while box center/size jitter is formulated as an issue of localization stability. Further, performing favorably in robotic scenarios, novel {\bf non-reference} assessments are proposed based on MOT rather than ground-truth labels. To this end, we modify MOT pipeline to capture recall failure (i.e., object missing) and design a Fourier approach for stability evaluation. In addition, including short tracklet suppression, fragment filling, and temporal location fusion, online tracklet refinement (OTR) is proposed to enhance VID continuity and stability. The proposed OTR can be generally applied to any detector in temporal tasks. Subsequently, we design small-overlap suppression (SOS) for extending VID approaches to SOT task, and thus SOT-by-detection is proposed. Compared to VID task, the SOS is able to induce a faster inference speed for SOT task. Compared to similarity-based SOT, the proposed SOT-by-detection has advantages on free initialization and flexibility. Our contributions are summarized as follows:
\begin{itemize}
  \item Two VID problems are novelly analyzed from the robotic perspective, i.e., continuity and stability, then non-reference assessments are proposed for them. Our assessments can make up the deficiency of traditional accuracy-based evaluation.
  \item We propose an OTR to generally improve detection continuity/stability. We also discuss how to solve these problems with the detector itself to inspire future works.
  \item We propose an SOS to extend VID approaches to the SOT task without the requirement of a similarity-based SOT module. The proposed SOT-by-detection is flexible for VID, MOT, and SOT tasks in robotic perception.
\end{itemize}

\section{Related Work}
\label{sec:RW}

%\subsection{Temporal Object Detection}
{\noindent\bf Temporal object detection.} There are manifold ideas of temporal detection, including 1) post-processing \cite{bib:SeqNms}, 2) tracking-based location \cite{bib:TCNN,bib:CDT,bib:DT,bib:DorT}, 3) feature aggregation \cite{bib:TSSD,bib:HPVD,bib:STSN}, 4) batch-frame processing \cite{bib:TPN}, 5) temporally sustained proposal \cite{bib:TRN}. All these ideas are attractive in that they can leverage temporal information for detection.

All above methods pursued high accuracy and followed AP evaluation. However, detection continuity and stability are equally important in robotic cases. AP considers static accuracy based on detection recall rate and precision \cite{bib:VOC}, but it can hardly give a temporal evaluation for VID methods as mentioned in Section~\ref{sec:intro}. Zhang and Wang proposed evaluation metrics for VID stability and proved that the stability has a low correlation with AP \cite{bib:Stable}. In detail, they formulated the stability problem as fragment error, center position error, and scale/ratio error. Their work was impressive but had two limitations: 1) they ignored the problem of short tracklet duration that was also an import situation in recall continuity; 2) their evaluation needed ground truth boxes, hampering their metrics from extensive applications. Conversely, we address these limitations and propose non-reference assessments without the need of man-made labels.

{\noindent\bf Tracking metrics.} Referring to \cite{bib:ClearMOT}, MOT metrics included multi-object tracking accuracy (MOTA) and precision (MOTP). MOTA synthesized false positive, false negative and identity switch of detected objects while MOTP considered the static localization precision. Therefore, tracklet fragment was captured by identity switch, but some other tracklet characteristics were ignored (e.g., short tracklet duration).

VOT used expect average overlap rate (EAO) for SOT evaluation \cite{bib:VOT}, including accuracy and robustness. Accuracy was determined by static IoU, and the robustness described tracking failure. After tracking failure captured by the evaluation process, the tracker would be initialized with ground truth. EAO could describe tracking fragments, but it could not give a comprehensive evaluation for multi-tracklets.

Tracking metrics (i.e., MOTA, MOTP, and EAO) are able to describe temporal recall, but similar to AP, they are insufficient for evaluating temporal localization. Moreover, all existing evaluations are based on ground truth. In contrast, we propose a Fourier approach to directly describe box jitter without the need of labels.

{\noindent\bf Tracking-by-detection (i.e., MOT).} MOT methods associate detected boxes across time. For example, Bewley \emph{et~al.} leveraged  Kalman and Hungarian methods for fast MOT \cite{bib:SORT}. Lu \emph{et~ al.} exploited RNN for sequential modeling and contrasted MOT approach with LSTM \cite{bib:ALSTM}. Almost all these methods are based on VID, but the effect of MOT on VID is usually ignored. Analytically, we report the MOT-based VID evaluation and enhancement. In addition, we exploit SOT-by-detection for flexible object perception.

{\noindent\bf Detection-SOT cascade.} Detection and SOT are distinct in their pipelines, and researchers tried to simultaneously leverage their advantages. Kang \emph{et~al.} used a tracking algorithm to re-score detection results with around tracklets \cite{bib:TCNN}. Kim~\emph{et~al.} combined a detector, a forward tracker, and a backward tracker to perform tracking-detection refinement \cite{bib:CDT}. Feichtenhofer \emph{et al.} simultaneously leveraged a two-stage detector and a correlation filter to boost VID accuracy \cite{bib:DT}. Luo \emph{et~al.} formulated detection and tracking as a sequential decision problem, and processed a frame by either a siamese tracker or a detector \cite{bib:DorT}. These methods complementally improved tracking and detection, but their SOT model and detector are independent so that high model complexity is usually incurred. Instead of the model cascade, we design an SOS to extend detection methods towards SOT task, producing SOT-by-detection framework.

\section{Approach}
\subsection{Non-Reference Assessments}
\label{sec:RC}
Our assessments follow a reasonable assumption, i.e., object motion is smooth across time without high-frequency location jitter or change of existence. We leverage a detector and an MOT module to recall all object tracklets in a video. In detail, any detector to be evaluated can be used for VID, and we employ the IoU-based MOT tracker reported by \cite{bib:TSSD} to associate detected boxes. Unlike label-based evaluation, we only focus on detected tracklets, because totally missed tracklet does not impact continuity and stability.

\begin{figure}[!t]
\begin{center}
\includegraphics[width=8cm]{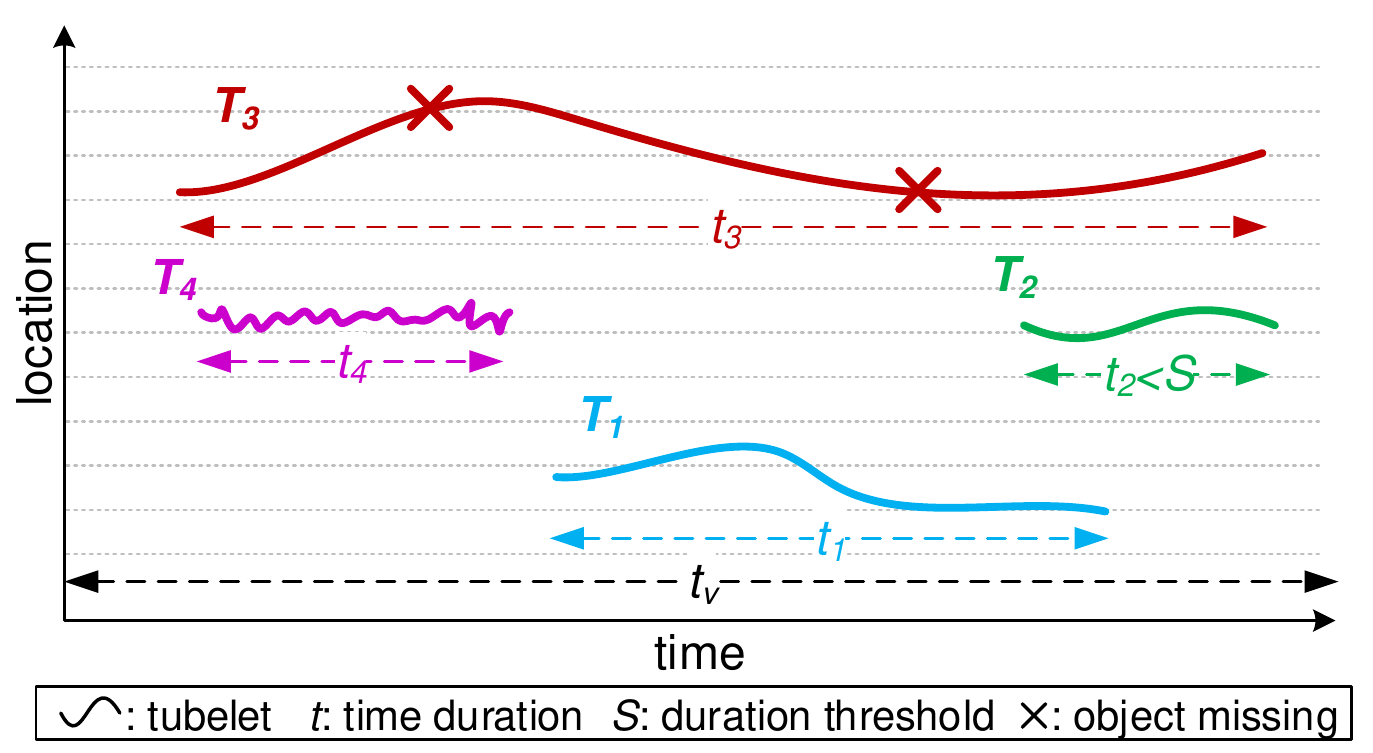}   % The printed column width is 8.4 cm.
\caption{Problem formation. Object tracklets could suffer from short tracklet duration (e.g., $\mathcal T_2$), fragments (e.g., $\mathcal T_3$), location jitter (e.g., $\mathcal T_4$).}% AP is insufficient for reflecting them.}
\label{fig:def}
\end{center}
\end{figure}

As delineated in Fig.~\ref{fig:def}, a detector locates and classifies objects at each frame $f$. Each object has confidence score $s$, box center $c_x, c_y$ and size $w, h$. $N$ tracklets $\{\mathcal T_n|{n=1,2,...,N}\}$ are produced after the whole video is processed by VID and MOT. The video duration and tracklet duration are denoted as $t_v, t_n$. The vertical axis of Fig.~\ref{fig:def} describes $c_x, c_y, w$, or~$h$.

%\subsubsection{Recall Continuity}
%\label{sec:RC}
{\noindent\bf Recall continuity.} Object tracklets should have an appropriate duration without interruption, and transient/intermittent object recall/missing damages recall continuity. As for this problem, we consider the impact of short tracklet duration and tracklet fragment. Referring to Fig.~\ref{fig:intro}~(a) and $\mathcal T_2$ in Fig.~\ref{fig:def}, tracklets with short duration frequently appear in VID. To capture them, we design extremely short duration error (ESDE) and short duration error (SDE) with various duration threshold as follows:
\begin{equation} \label{eqn:SDE}
\text{ESDE/SDE} = \frac{1}{t_v}\sum_{n=1}^N
 (t_n \quad \text{if} \quad t_n<S \quad \text{else} \quad 0).
\end{equation}
If $\mathcal B$ is true, then $(\mathcal A \quad \text{if} \quad \mathcal B \quad \text{else} \quad \mathcal C)$ returns $\mathcal A$; otherwise, it returns $\mathcal C$. In this paper, $S_{ESDE}=3,S_{SDE}=10$, which describe different degrees of short duration problem.

Fig.~\ref{fig:intro}~(a) and $\mathcal T_3$ in Fig.~\ref{fig:def} describe tracklet fragment problem. Some MOT algorithms end a tracklet after a recall failure. Conversely, we count the number of continuous recall failure with $S_{lost}$, and leave a $S^{max}_{lost}$-frame life duration for each tracklet. That is, if a recall-failed tracklet is re-matched by a box in consequent $S^{max}_{lost}$ frames, the tracklet can be kept. In this way, the total number of object missing $om$ in the whole tracklet can be captured, forming tracklet fragment error (TFE) and fragmental tracklet ratio (FTR) as follows:
\begin{equation} \label{eqn:TFE}
%\begin{array}l
\setlength{\jot}{2pt}
\begin{aligned}
&\text{TFE} = \sum_{n=1}^N om_n/\sum_{n=1}^Nt_n \\
&\text{FTR} = \frac{1}{N}\sum_{n=1}^N(1 \quad \text{if} \quad om_n>0 \quad \text{else} \quad 0)
\end{aligned}.
%\end{array}.
\end{equation}

TFE describes the ratio of object missing in tracklets, while FTR gives the ratio of faulty tracklets. They are complementary for tracklet fragment problem, i.e., a better VID result should have lower TFE and FTR in the meantime. That is, there is a small number of object missing, and object missing is concentrated in a small number of tracklets. Note that these calculations are numerically small, so a log transformation is used to enhanced contrast, i.e., $\log_{100}(1+99\times\alpha)$, where $\alpha$ represents ESDE, SDE, TFE, or FTR. Finally, recall continuity error (RCE) is defined as $\text{RCE}=\text{ESDE}+\text{SDE}+\text{TFE}+\text{FTR}$.

%\subsubsection{Localization Stability}
{\noindent\bf Localization stability.} Object tracklets should be smooth in localization, and box center/size jitter damages localization stability (see Fig.~\ref{fig:intro} (b) and $\mathcal T_4$ in Fig.~\ref{fig:def}). We evaluate temporal stability in Fourier domain so that our approach can work without labels. Time domain data $p$ can be transformed to Fourier domain by $P=\mathcal F (p)$, where $p$ represents $c_x,c_y,w$, or $h$. Thus, $P$ contains frequency information of $p$, and we extract frequency-related amplitude with $\widetilde P=\text{Abs}(P)$. Note that each tracklet produces different frequency component because of variable data length (i.e., tracklet duration). That is, $\widetilde P=\{(q^p_k,A^p_k)\},q^p_k=k/t,k=\{0,1,...,\lfloor t/2\rfloor\}$. Here, $q$ is the frequency set; $t$ is tracklet duration; and $A$ denotes frequency-related amplitude. Based on Fourier analysis, center jitter error (CJE) and size jitter error (SJE) are designed~as
\begin{equation} \label{eqn:JE}
%\begin{array}l
\setlength{\jot}{2pt}
\begin{aligned}
&\text{CJE} = (10^3\sum_{n=1}^N\sum_{p\in\{c_x,c_y\}}\sum_{k=1}^{\lfloor t_n/2\rfloor} q^p_{n,k}A^p_{n,k})/\sum_{n=1}^Nt_n \\
&\text{SJE} = (10^3\sum_{n=1}^N\sum_{p\in\{w,h\}}\sum_{k=1}^{\lfloor t_n/2\rfloor} q^p_{n,k}A^p_{n,k})/\sum_{n=1}^Nt_n \\
%&\text{LJE} = \text{CJE} + \text{SJE} \\
\end{aligned},
%\end{array}.
\end{equation}
%where $k$ starts from 1 since $q^p_{n,0}=0$.
Ultimately, localization jitter error $\text{LJE} = \text{CJE} + \text{SJE}$.
%Object size is usually described by scale (i.e., $wh$) and aspect ratio (i.e., $w/h$). However, SJE directly analyzes $w$ and $h$, which has an equal ability to describe object size. Moreover, our decoupled \text{SJE} is able to definitely describe stability in both directions.

\subsection{Online Tracklet Refinement}
\label{sec:VIDMOT}
For enhancing recall continuity and localization stability, we refine VID results based on tracklets. A new attribute is used to describe tracklets, i.e., current duration $S_{dur}$. Therefore, a tracklet can be formulated as $\mathcal T=(\mathcal D,ID,S_{lost},S_{dur})$, where $S_{dur}$ records tracklet duration at each timestamp; $S_{lost}$ has been explained in Section \ref{sec:RC}; $ID$ denotes tracklet identity; $\mathcal D$ is the object set in the tracklet (i.e., $\{(s,c_x,c_y,w,h)\}$), and the length of $\mathcal D$ (i.e.,~$S_{obj}$) cannot exceed $S^{max}_{obj}=5$. That is, if $S_{dur}>S^{max}_{obj}$, only the latest $S^{max}_{obj}$ objects are preserved in $\mathcal D$.

{\noindent\bf Short tracklet suppression.} For suppressing short tracklets and enhancing ESDE/SDE, we define a tracklet as reliable tracklet if $S_{dur}>S_{SDE}$, then boxes in unreliable tracklets are suppressed. This manner is beneficial to continuity, and it has two-fold effects on accuracy. Firstly, false positives could be suppressed, because their recall is usually inconsecutive across time so that a reliable tracklet is hard to form. Secondly, false negatives could be produced since an object would not be reported until it forms an $S_{SDE}$-length tracklet.

{\noindent\bf Fragment filling.} In terms of the fragment issue and TFE/FTR, we make up object missing in a tracklet based on a reasonable assumption, i.e.,  the object motion is uniform in an extremely short duration (e.g., $S^{max}_{obj}$). When a tracklet suffers from a recall failure at the $f$th frame, its previous boxes $\{(c^{f-i}_x,c^{f-i}_y,w^{f-i},h^{f-i})| {i=1,2,...,S_{obj}}\}$ can be used to predict current location. In detail, we first estimate the velocity $v_p$, i.e., $v_p=(\sum_{i=1}^{S_{obj}-1}p^{f-i}-p^{f-i-1})/(S_{obj}-1)$, then the current location can be given as $p=p^{f-1}+v_p$, where $p$ denotes $c_x,c_y,w$, or $h$.

{\noindent\bf Temporal location fusion.} For location stability and CJE/SJE, we add the object into its tracklet, then produce a new location with weighted average. A geometric progression is contrasted with $\Omega=\{\omega^l | l=1,...,0.1\}$, where $l$ is an $S_{obj}$-length arithmetic progression. The normalized $\Omega$ is utilized as the weight to merge $\{p|p\in \mathcal D \& \mathcal D\in \mathcal T\}$, and the updated location can be formulated as $\hat p^f=\sum_{i=0}^{S_{obj}}\Omega_{i}p^{f-i}$.

\subsection{SOT-by-Detection}

\begin{algorithm}[!t]
\label{alg:sos}
        \caption{SOS-NMS}
        \KwIn{After selection by confidence threshold, boxes $\mathcal B=\{b_1,...,b_m\}$, confidence scores $\mathcal S=\{s_1,...,s_m\}$; previous tracked box $b^{f-1}$; SOS/NMS thresholds $U^{sos},U^{nms}$}
        \KwOut{Tracked box $b^{f}$}
        \Begin{
            \tcp{SOS based on IoU ("+=/-=" denotes element add/removal)}
            $\mathcal B^{sos}= \mathcal B; \mathcal S^{sos}= \mathcal S; \mathcal O^{sos}= iou(b^{f-1},\mathcal B)$ \\
            \For{$(b_i,s_i, o_i)\in (\mathcal B^{sos},\mathcal S^{sos},\mathcal IoU^{sos})$ }{
                \If{$o_i<U^{sos}$}
                    {$\mathcal B^{sos}-=b_i; \mathcal S^{sos}-=s_i; \mathcal O^{sos}-=o_i$}
            }
            \tcp{Inspection of tracking failure}
            \If{$\mathcal B^{sos} = empty$}
                {return $b^{f}= empty$}
            \tcp{NMS based on confidence score}
            $\mathcal B^{nms}=\{\}; \mathcal S^{nms}=\{\}; \mathcal O^{nms}= \{\}$ \\
            \While{$\mathcal B^{sos}\ne empty$}{
                $idx=\text{argmax} \mathcal S^{sos}$ \\
                $b=\mathcal B^{sos}_{idx}; s=\mathcal S^{sos}_{idx}; o=\mathcal O^{sos}_{idx}$ \\
                $\mathcal B^{nms}+=b; \mathcal S^{nms}+=s; \mathcal O^{nms}+=o$ \\
                $\mathcal B^{sos}-=b; \mathcal S^{sos}-=s; \mathcal O^{sos}-=o$ \\
                \For{$(b_i,s_i)\in (\mathcal B^{sos},\mathcal S^{sos})$ }{
                    \If{$iou(b,b_i)>U^{nms}$}
                        {$\mathcal B^{sos}-=b_i; \mathcal S^{sos}-=s_i; \mathcal O^{sos}-=o_i$}
                }
            }
            \tcp{Selection of single box with IoU}
            $idx=\text{argmax} \mathcal O^{nms}$ \\
            return $b^{f}=\mathcal B^{nms}_{idx}$
        }
\end{algorithm}

%\subsubsection{Small-Overlap Suppression}
{\noindent\bf Small-overlap suppression.} We promote VID model to generate SOT result by propagating previous location $b^{f-1}=(c^{f-1}_x,c^{f-1}_y,w^{f-1},h^{f-1})$ before non-maximum suppression (NMS). Taking inspiration from NMS, we leverage IoU-based suppression to this end. Referring to Algorithm~\ref{alg:sos}, after selection by confidence threshold, IoU between candidate boxes and $b^{f-1}$ is calculated, then candidate boxes with small IoU (e.g.,~$<U^{sos}$) are discarded. Next, tracking failure would be reported if all boxes are suppressed by SOS. Subsequently, NMS is performed on the remaining boxes. Finally, we select a box with IoU maximum as current SOT result $b^f$. Compared to the manner of IoU-based re-scoring, the SOS does not affect confidence scores. In our opinion, confidence score and IoU are two different properties of objects, where confidence score describes object category while IoU reports object motion. Thereby, the SOS-NMS is based on alternating confidence score and IoU, i.e., 1) discarding obviously category-incorrect candidates with confidence threshold; 2) discarding obviously motion-incorrect candidates using IoU threshold; 3) discarding candidates without local maximum of confidence score; 4) generating single object location with IoU maximum. Note that SOS-NMS has a speed advantage over NMS since a significant amount of candidate boxes are suppressed by computationally efficient SOS.

\begin{figure}[!t]
\begin{center}
\includegraphics[width=7cm]{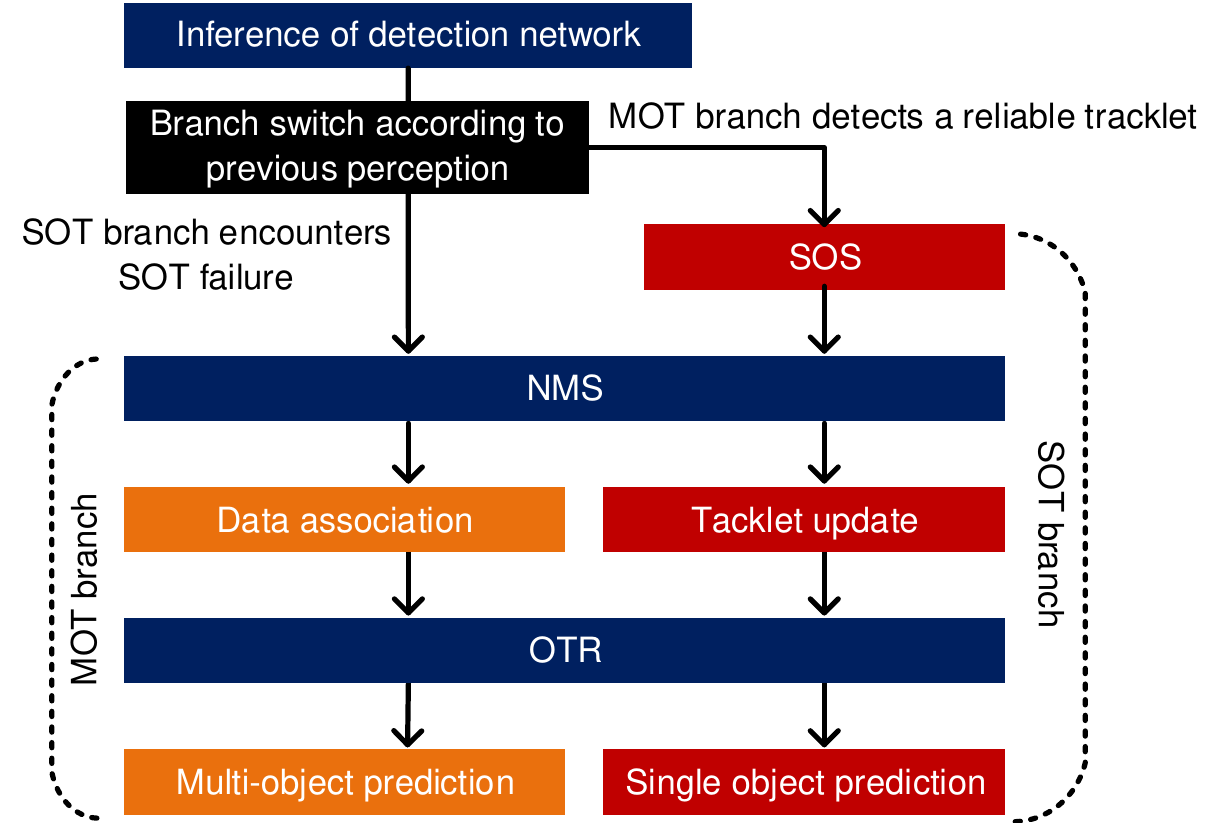}   % The printed column width is 8.4 cm.
\caption{SOT-by-detection with our proposed SOS and OTR. The MOT branch is designed to search initial box for SOT. Reliable tracklet is captured by OTR, while SOT failure is captured by SOS. If no switch condition is met, the previous behavior is continuously performed.}
\label{fig:MOTSOT}
\end{center}
\end{figure}

\begin{figure*}[!t] \centering
\subfigure { \label{fig:img_seq}
\includegraphics[width=16cm]{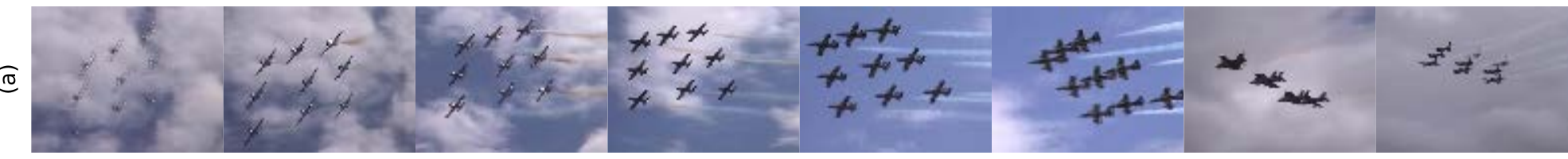}
}
\subfigure { \label{fig:ssd_plot}
\includegraphics[width=16cm]{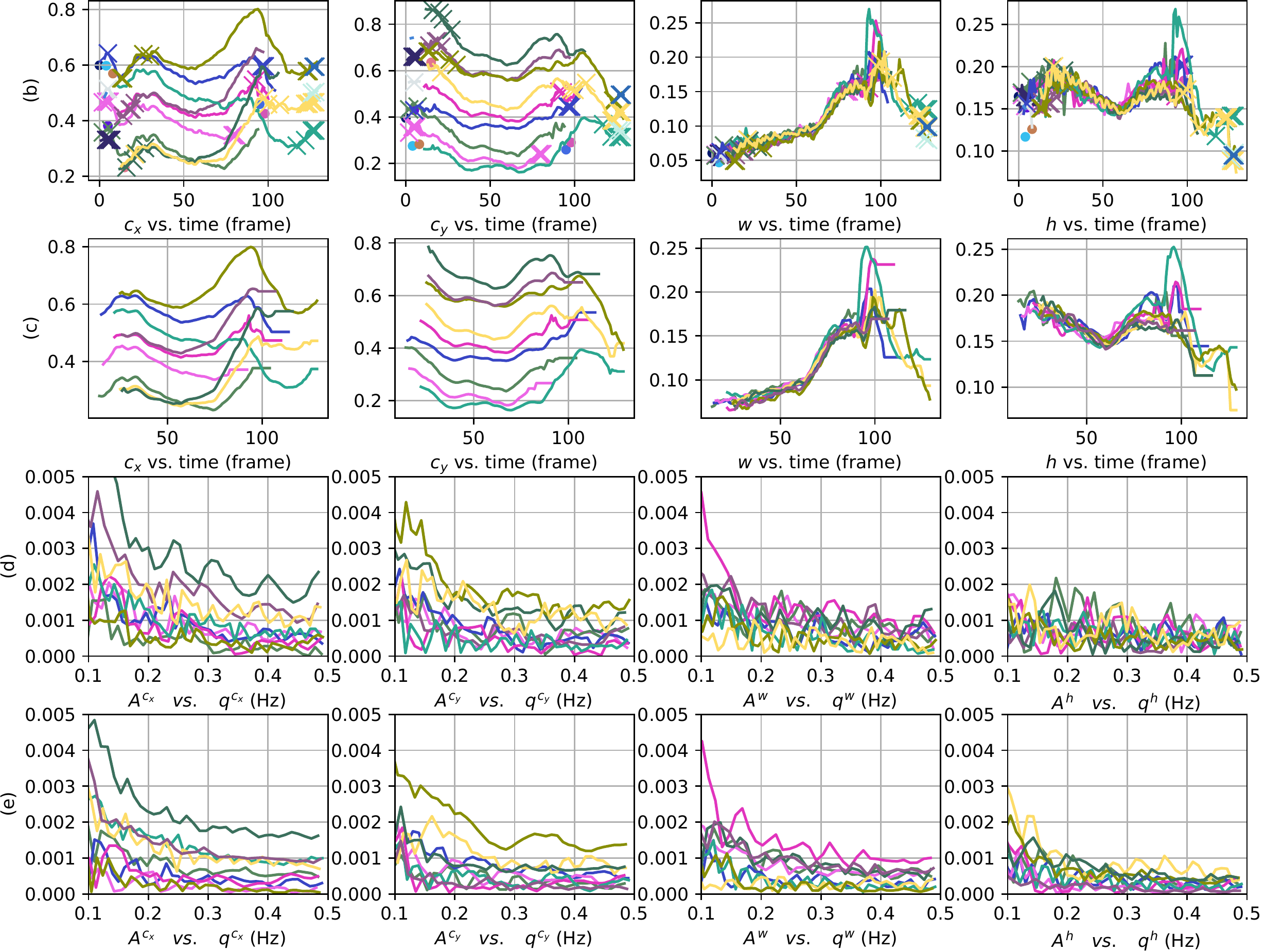}
}
\caption{Tracklet visualization for SSD. (a) The video snippet; (b) original tracklets; (c) refined tracklets with OTR; (d) original Fourier results; (e) Fourier results with OTR. Colors differentiate $ID$ and ``$\times$'' denotes object missing at a time-stamp. $A$ and $q$ are defined in Section~\ref{sec:RC}).} %The unit of the horizontal axis of (b,c) is frame while that of (d,e) is Hz. $A$ and $q$ are defined in Section~\ref{sec:RC}).}
\label{fig:plot}
\end{figure*}

%\subsubsection{Unified VID-MOT-SOT Framework}
{\noindent\bf SOT-by-detection framework.} As shown in Fig.~\ref{fig:MOTSOT}, the proposed SOT-by-detection framework only adopts a detection model, and an MOT branch is developed to search initial object location for SOT. We first define the condition of MOT-SOT switch: 1) MOT is initially performed; 2) when a reliable tracklet (i.e., $S_{dur}>S_{SDE}$ captured by OTR) is found, the SOT branch is activated to track this reliable tracklet. If there are several reliable tracklets, only one would be tracked according to confidence score; 3) the MOT branch is re-activated after SOT failure captured by SOS. In terms of components, the MOT branch includes NMS, data association, and OTR; the SOT branch contains SOS, NMS, tracklet update, and OTR. For SOT, OTR only processes the tracked tracklet. Remarkably, the SOT branch is faster than the MOT branch becasue 1) SOS is able to highly reduce NMS's computational cost; 2) data association in the MOT branch is usually time-consuming. The proposed SOT-by-detection has three-fold advantages in SOT task: 1) There is no need for man-made initial location; 2) confidence score is more reliable than similarity score because of semantic classification, so SOT-by-detection can capture tracker state (e.g., tracking drift or failure) in a more reliable manner; 3) complex detection-tracking cascade is avoid. Furthermore, our proposed framework can be easily extended to 2-object tracking, 3-object tracking, and so on.

\begin{table*}[!t]
\renewcommand{\arraystretch}{1.0}
\setlength\tabcolsep{10pt}
\caption{Continuity and stability evaluation of several existing detectors based on the proposed non-reference metrics.}
\label{tab:metric}
\centering
\begin{tabular}{c | c | c c c c c | c c c}
\Xhline{1.5pt}
\multirow{ 2}{*}{Method}      & \multirow{ 2}{*}{mAP} & \multicolumn{5}{c|}{Recall continuity} & \multicolumn{3}{c}{Localization stability ($\omega=10$)} \\
                              &         & ESDE  & SDE   & TFE   & FTR   & RCE  & CJE   & SJE & LJE\\
\hline
\emph{w/o OTR}                &         &       &       &       &       &      &       &  \\
\emph{static methods}         &         &       &       &       &       &      &       &         \\
SSD \cite{bib:SSD}            & 0.630   & 0.062 & 0.234 & 0.320 & 0.246 & 0.862& 0.242 & 0.334 & 0.576 \\
RetinaNet \cite{bib:RetinaNet}& 0.656   & 0.060 & 0.250 & 0.350 & 0.283 & 0.943& 0.236 & 0.317 & 0.553\\
RefineDet \cite{bib:RefineDet}& 0.669   & 0.126 & 0.350 & 0.391 & 0.306 & 1.173& 0.257 & 0.362 & 0.619\\
DRN \cite{bib:DRN}            & \bf0.694& 0.114 & 0.330 & 0.389 & 0.312 & 1.145& 0.248 & 0.346 & 0.594\\
\hline
\emph{temporal methods}       &         &       &       &       &       &      &  \\
TRN \cite{bib:TRN}            & 0.665   & 0.120 & 0.334 & 0.375 & 0.265 & 1.094& 0.252 & 0.346 & 0.598 \\
TDRN \cite{bib:TRN}           & 0.673   & 0.116 & 0.345 & 0.388 & 0.297 & 1.146& 0.247 & 0.360 & 0.607 \\
TSSD \cite{bib:TSSD}          & 0.654   &\bf0.059&\bf0.206&\bf0.257&\bf0.240&\bf0.762&\bf 0.210&\bf0.253 &\bf 0.463 \\
\hline
\hline
\emph{w/ OTR}                 &         &       &       &       &       &      &       &            \\
SSD                           &  $-$    & 0.003 & 0.026 & 0.0   & 0.0   & 0.029& 0.169 & 0.208 & 0.377 \\
RetinaNet                     &  $-$    & 0.003 & 0.023 & 0.0   & 0.0   & 0.026& 0.168 & 0.204 & 0.372 \\
RefineDet                     &  $-$    & 0.004 & 0.037 & 0.0   & 0.0   & 0.041& 0.173 & 0.212 & 0.385 \\
DRN                           &  $-$    & 0.003 & 0.036 & 0.0   & 0.0   & 0.039& 0.172 & 0.208 & 0.380 \\
\hline
TRN                           &  $-$    & 0.003 & 0.030 & 0.0   & 0.0   & 0.033& 0.171 & 0.209 & 0.380 \\
TDRN                          &  $-$    & 0.004 & 0.031 & 0.0   & 0.0   & 0.035& 0.170 & 0.218 & 0.388 \\
TSSD                          &  $-$    & 0.003 & 0.029 & 0.0   & 0.0   & 0.032& 0.159 & 0.180 & 0.339 \\
\Xhline{1.5pt}
\end{tabular}
\end{table*}

\section{Experiments and Analysis}
{\noindent\bf Preliminary.} We analyze real-time online detectors, i.e., SSD \cite{bib:SSD}, RetinaNet \cite{bib:RetinaNet}, RefineDet \cite{bib:RefineDet}, DRN \cite{bib:DRN}, TSSD \cite{bib:TSSD}, TRN, and TDRN \cite{bib:TRN}. The first four are static detectors while the last three are temporal methods. These detectors have close relation and inheritance, so we unveil the effects of their designs on temporal performance based on our metrics. In return, these evaluations can verify the effectiveness of our assessments. SSD detects objects in a single-stage manner \cite{bib:SSD}. Based on SSD, RetinaNet adopts feature pyramid networks (FPN) to enhance shadow-layer receptive field \cite{bib:RetinaNet}. Based on RetinaNet, RefineDet introduces a two-step regression to the single-stage pipeline \cite{bib:RefineDet}. Based on RefineDet, DRN performs joint anchor-feature refinement for detection \cite{bib:DRN}. Referring to Section~\ref{sec:RW}, there are 5 types of VID approaches, but post-processing and tracking-based methods actually adopt static detectors, and batch-frame approaches can hardly work in real-world scenes, so we analyze the methods with feature aggregation or temporally sustained proposal. Based on SSD, TSSD uses attentional-LSTM for aggregating visual features across time \cite{bib:TSSD}. As temporally sustained proposal approaches, TRN and TDRN propagate refined anchors and feature offsets across time based on RefineDet and DRN \cite{bib:TRN}. All these detectors are trained and evaluated on ImageNet VID dataset \cite{bib:VID}. Both confidence threshold and NMS threshold are fixed as $0.5$. Ultimately, the merit of SOT-by-detection with TDRN as the detector is verified in a real-world robotic grasping task.

\subsection{Analysis on VID Continuity/Stability}

%\subsubsection{Visualization of Detection Tracklets}
{\noindent\bf Tracklet visualization.} As shown in Fig.~\ref{fig:plot}~(a), we use a VID case with 9 object instances to visualize SSD detection. Referring to Fig.~\ref{fig:plot}~(b), SSD suffers from serious continuity and stability problems. At the beginning of this video, a vast number of object missing (i.e., $\times$ on curves) and short tracklets (i.e., short curves and scattered points) appear due to motion blur. Then, continuity problems appear again at the end of the video owing to occlusion. For localization stability, Fig.~\ref{fig:plot}~(d) plots the amplitude of high-frequency component ($>0.1$~Hz) in Fourier domain. Numerically, $\text{ESDE}=0.448, \text{SDE}=0.715, \text{TFE=}0.410, \text{FTR}=0.619, \text{RCE}=2.192, \text{CJE}=0.264, \text{SJE}=0.183, \text{LJE}=0.447$.

OTR is able to refine SSD results from the perspective of tracklet. As shown in Fig.~\ref{fig:plot}~(c), OTR eliminates all short tracklets and fragments, and refined tracklets are smoother. Referring to Fig.~\ref{fig:plot}~(e), high-frequency amplitude in Fourier domain is suppressed to some degree. As a result, $\text{ESDE}=0.0, \text{SDE}=0.0, \text{TFE}=0.0, \text{FTR}=0.0, \text{RCE}=0.0, \text{CJE}=0.219, \text{SJE}=0.142, \text{LJE}=0.361$.

%\subsubsection{Numerical Evaluation}
{\noindent\bf Numerical evaluation.} Referring to Table~\ref{tab:metric}, detectors are evaluated with our proposed non-reference assessments on VID validation set. The accuracy-best method is DRN with $69.4\%$ mAP. However, there is low correlation between accuracy and continuity/stability. From static SSD and RetinaNet, we observe that FPN improves localization stability since spatial feature fusion. However, as for continuity, RetinaNet performs worse than SSD since more hard objects can be detected by RetinaNet. That is, detecting hard objects (e.g., small objects) can easily produce continuity problems, and SSD is likely to completely miss them because of relatively low detection accuracy. Besides, the comparison between RefineDet and RetinaNet indicates that anchor refinement improves accuracy but induces more serious problems on continuity and stability, because anchor-feature mis-alignment is exacerbated. Finally, DRN conducts joint anchor-feature refinement to relieve RefineDet's drawback, i.e., features are relatively accurate for describing refined anchors. Thus, DRN performs better than RefineDet on almost all metrics.

\begin{figure}[!t]
\begin{center}
\includegraphics[width=6.5cm]{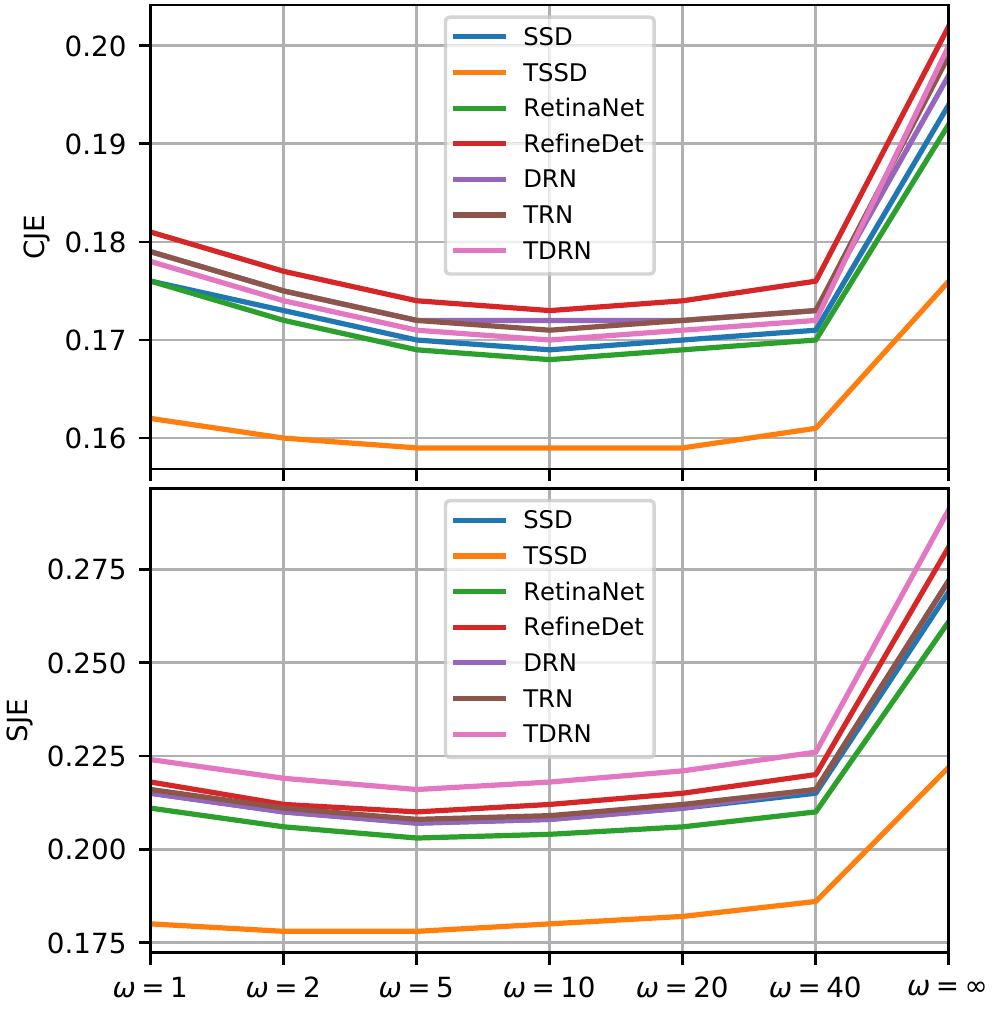}   % The printed column width is 8.4 cm.
\caption{Plot of CJE/SJE vs. $\omega$. $\omega=\infty$ indicates that temporal location fusion loses its effect.}
\label{fig:w_cje_sje}
\end{center}
\end{figure}

For temporal approaches, we draw readers' attention to three detector pairs, i.e., TRN vs. RefineDet, TDRN vs. DRN, and TSSD vs. SSD, where the temporal detector is extended from the static one. For TRN vs. RefineDet and TDRN vs. DRN, temporal detectors obtain on par, sometimes even worse, performance compared with static approaches. Therefore, the design of temporally sustained proposal has an ignorable effect on detection continuity/stability. In contrast, TSSD performs better than SSD by a large margin on all metrics, which validates the effectiveness of temporal feature aggregation. That is, TSSD can smooth visual features across time, and produces more temporally consistent results.

Based on OTR, all metrics can be effectively improved for all tested approaches. OTR can totally eliminate fragment problem by filling up recall failure, generating $0$ TFE/FTR. In terms of ESDE, SDE, CJE, and SJE, OTR also produces substantially better results. Note that AP cannot be reported with OTR, because MOT needs a relatively high confidence threshold (e.g., $0.5$) for data association while AP evaluation usually uses a low threshold (i.e., $0.01$).

We use $\Omega=\{\omega^l|{l=1,...,0.1}\}$ to fuse current prediction with location history, so $\omega$ controls the temporal location fusion. We investigate $\omega=1,2,5,10,20,40$, which induces decay ratios of $1,1.17,1.44,1.68,1.96,2.29$ across time. That is, small $\omega$ merges location information with roughly equal weights, while large $\omega$ produces greater weights for recent timestamps. Note that $\omega=\infty$ means temporal location fusion loses its effect. For example, $\Omega=\{0.437,0.260,0.155,0.092,0.055\}$ when $\omega=10,S_{obj}=5$. Referring to Fig.~\ref{fig:w_cje_sje}, optimal $\omega$ ranges from $5$ to $10$, where fusion ratio is suitable for localization~stability.

\subsection{SOT-by-Detection}
%\subsubsection{Comparison of NMS and SOS-NMS on Speed}
\begin{figure}[!t]
\begin{center}
\includegraphics[width=7cm]{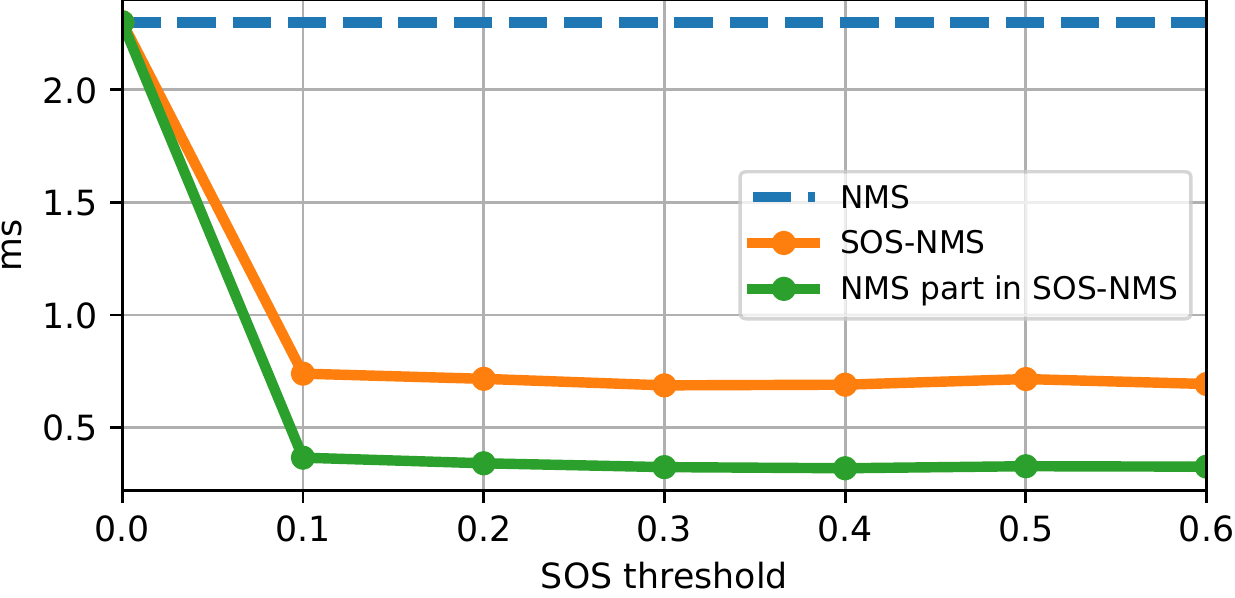}   % The printed column width is 8.4 cm.
\caption{Plot of time consumption of NMS and SOS-NMS.}
\label{fig:SOS_time}
\end{center}
\end{figure}

{\noindent\bf Speed comparison of NMS and SOS-NMS.} This test is conducted with an Intel 2.20 GHz Xeon(R) E5-2630 CPU. As plotted in Fig.~\ref{fig:SOS_time}, the SOS considerably reduces the time consumption of candidate selection. NMS takes $2.30$~ms per~frame, and SOS can highly reduce the box amount for NMS with temporal information. As a result, SOS-NMS's time cost can be reduced to $0.69$~ms~per~frame, and the NMS part in SOS-NMS only takes $0.33$~ms~per~frame when SOS threshold $\ge0.3$. Thereby, when performing SOT based on SOS, the VID model can achieve faster speed. In the following experiments, SOS threshold is fixed as $0.3$.

%\subsubsection{Siamese SOT vs. Unified VID-MOT-SOT}
\begin{figure}[!t]
\begin{center}
\includegraphics[width=8cm]{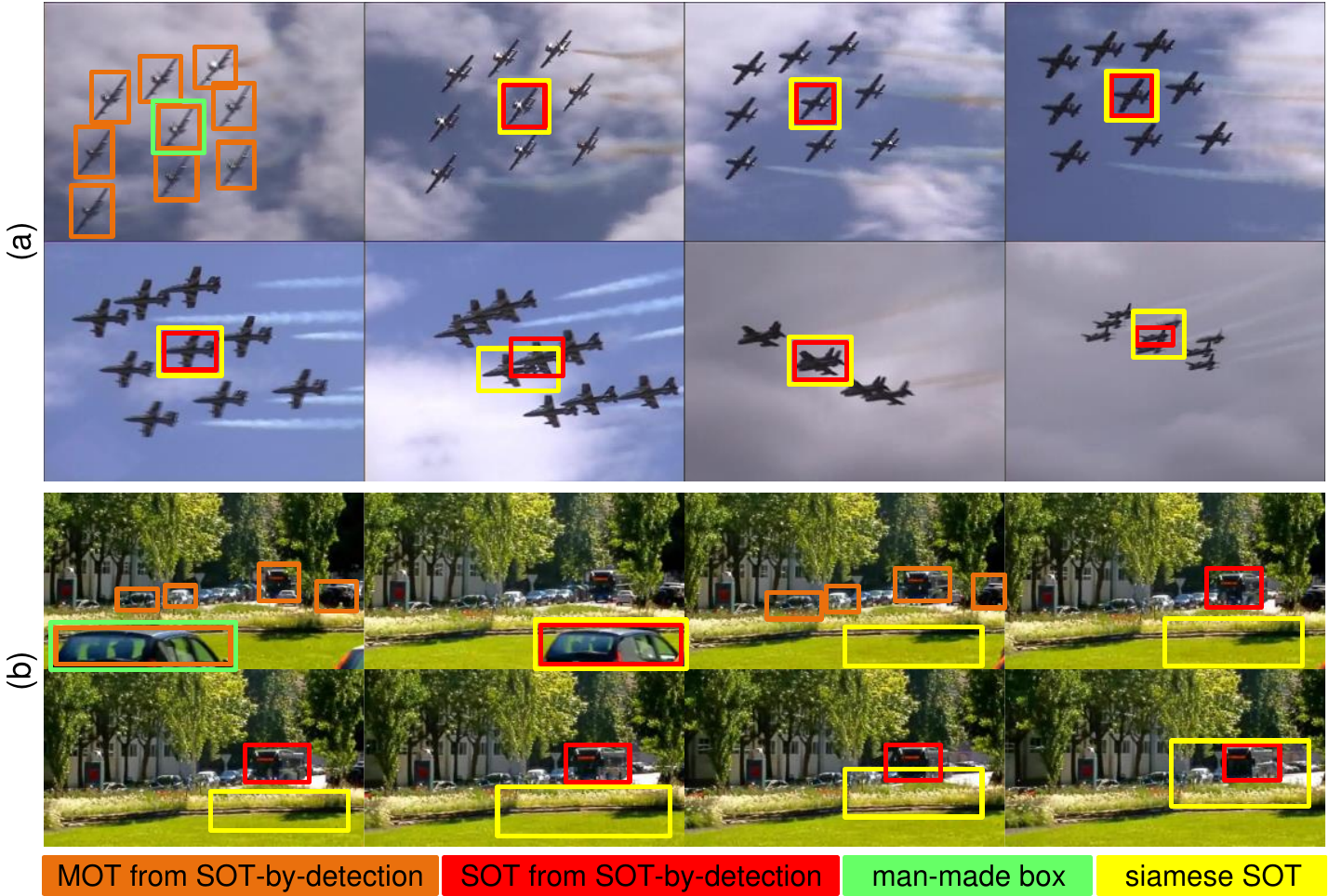}   % The printed column width is 8.4 cm.
\caption{Comparison between SOT-by-detection and siamese SOT. If tow boxes are highly overlapping, we slightly change their sizes for visualization.}
\label{fig:SOTcompare}
\end{center}
\end{figure}

{\noindent\bf SOT-by-detection vs. siamese SOT.} There lacks a dataset to quantitatively evaluate SOT-by-detection and siamese SOT. That is, siamese SOT cannot work on VID dataset (e.g., ImageNet VID \cite{bib:VID}) because it cannot predict object category, and VID model in SOT-by-detection cannot be trained with a category-free SOT dataset (e.g., VOT \cite{bib:VOT}). Thus, we qualitatively compare our method and a siamese tracker \cite{bib:SiamFC++} on ImageNet VID dataset. Firstly, SOT-by-detection has the ability to search objects, i.e., MOT can provide an initial localization for SOT. Then, siamese SOT is particularly susceptible to unconscious tracking drift (as delineated in Fig.~\ref{fig:SOTcompare}~(a)). Finally, referring to Fig.~\ref{fig:SOTcompare}~(b), SOT-by-detection is able to capture tracking failure and conveniently re-activate the MOT branch for object search (see the 3rd snapshot). On the contrary, the siamese tracker always reports a similar region, and it is unable to start a second tracking.

\begin{figure}[!t]
\begin{center}
\includegraphics[width=8cm]{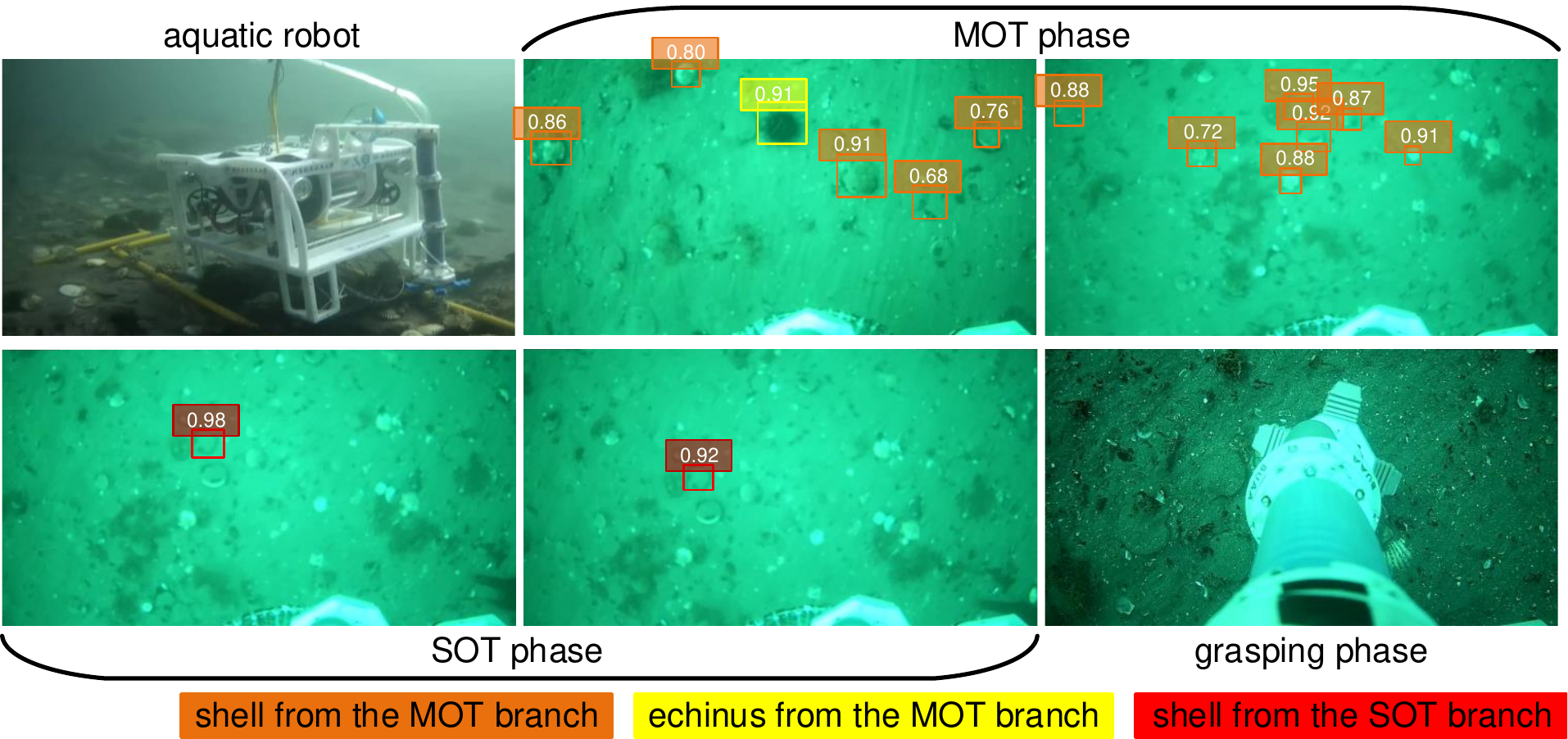}   % The printed column width is 8.4 cm.
\caption{SOT-by-detection in object grasping task. Based on an aquatic robot, SOT-by-detection can provide robust visual information with flexible detection and tracking for robotic navigation and grasping.}
\label{fig:robo}
\end{center}
\end{figure}

{\noindent\bf SOT-by-detection in robotics.} This framework is potential and flexible in robotic perception. For example, when a mobile robot tries to grasp all objects in an area, the MOT branch should firstly work for object search, i.e., perception of object group. After a reliable tracklet is detected, the SOT branch should work for elaborate perception of object instance. At this time, other perception manners (e.g., mask, depth, etc.) can be added for more object instance information to guide this object grasping. If the target encounters tracking failure, our proposed framework can immediately capture this situation and give a convenient way to switch to the MOT branch for object search.

As shown in Fig.~\ref{fig:robo}, an autonomous aquatic robot is developed for grasping marine products (e.g., echinus and shell) on nature seabed. The test venue is located in Zhangzidao, China. In this task,  SOT-by-detection is competent in detecting/tracking objects for robotic perception, and provide flexible perception ability for object group/instance. Based on SOT-by-detection, our robot is able to efficiently approach and grasp targets. The experimental video is available at \url{https://youtu.be/9HG4JagAUpw}.

\subsection{Discussions}

%\subsubsection{Improvement of Continuity and Stability from the Detector Itself}
{\noindent\bf Detector-based improvement.} This paper evaluates and enhances detection continuity and stability from the perspective of MOT. Additionally, our evaluation indicates that temporal performance is benefited from feature aggregation. On one hand, spatial/scale smooth is effective (see RetinaNet vs. SSD), and on the other hand, temporal fusion is more efficient (see TSSD vs. SSD). Thus, we advocate investigating fusion approaches for improving continuity/stability from the detector itself. For example, Besides RNN in TSSD, Zhu \emph{et~al.} aggregated temporal features with flow-based motion estimation \cite{bib:HPVD}, and Bertasius~\emph{et~al.} leveraged the deformable convolutions to constructed robust temporal features \cite{bib:STSN}.

%\subsubsection{Similarity-based SOT vs. VID-based SOT}
{\noindent\bf Limitation of SOT-by-detection.} SOT-by-detection has advantages but it also has a limitation that it cannot deal with unseen categories or object part. On one hand, similarity-based SOT and SOT-by-detection are complementary. That is, the similarity module focus on appearance similarity while SOT-by-detection is dependent on category attribute and motion information. Therefore, they can be reasonably combined for addressing this problem. On the other hand, we advocate solving this limitation with a combination of online learning \cite{bib:ATOM} and few-short learning \cite{bib:FewShot}.

\section{Conclusion}
This paper has dealt with detection continuity/stability and SOT-by-detection for robotic perception. Firstly, we analyze that AP has difficulty in reflecting temporal continuity/stability, and propose non-reference assessments to evaluate them. The evaluation of recall continuity is based on tracklet analysis and the localization stability problem is captured by the Fourier method. Secondly, we design OTR to enhance detection continuity/stability by short tracklet suppression, fragment filling, and temporal location fusion. Finally, SOS is proposed to extend VID methods towards SOT task, and the SOT-by-detection is developed. With ImageNet VID dataset, the effects of our methods are verified. As a result, our non-reference assessments and OTR can effectively deal with temporal performance in VID, and SOT-by-detection plays an essential role in robotic perception.

We will enhance temporal performance with feature fusion and improve SOT-by-detection by online learning.


\begin{thebibliography}{99}

\bibitem{bib:VID}
O.~Russakovsky, J.~Deng, H.~Su, J.~Krause, S.~Satheesh, S.~Ma, Z.~Huang, A.~Karpathy, A.~Khosla, M.~Bernstein, A.~C.~Berg, and F.~Li, ``ImageNet large scale visual recognition challenge,'' \emph{Int. J. Comput. Vis.}, vol. 115, no. 3, pp. 211--252, 2015.

\bibitem{bib:MOT}
L.~Leal-Taixe, A.~Milan, I.~Reid, S.~Roth, and K.~Schindler, ``Motchallenge 2015: Towards a benchmark for multi-target tracking,'' \emph{arXiv:1504.01942}, 2015.

\bibitem{bib:VOT}
M.~Kristan, et al., ``The sixth visual object tracking vot-2018 challenge results,'' in \emph{Proc. Eur. Conf. Comput. Vis. workshops}, Munich, Germany, Sept. 2018.

\bibitem{bib:VideoTub}
K.~Kang, W.~Ouyang, H.~Li, X.~Wang, ``Object detection from video tubelets with convolutional neural networks,'' in \emph{Proc. IEEE Conf. Comput. Vis. Pattern Recognition}, Las Vegas, the US, Jun. 2016, pp. 817--825.

\bibitem{bib:SeqNms}
W.~Han, P.~Khorrami, T.~L.~Paine, P.~Ramachandran, M.~Babaeizadeh, H.~Shi, J.~Li, S.~Yan, and T.~S.~Huang, ``Seq-NMS for video object detection,'' \emph{arXiv:1602.08465}, 2016.

\bibitem{bib:DT}
C.~Feichtenhofer, A.~Pinz, A.~Zisserman, ``Detect to track and track to detect,'' in \emph{Proc. IEEE Conf. Comput. Vis. and Pattern Recognition}, Venice, Italy, Oct. 2017, pp. 3038--3046.

\bibitem{bib:DorT}
H.~Luo, W.~Xie, X.~Wang, W.~Zeng, ``Detect or track: Towards cost-effective video object detection/tracking,'' in \emph{Proc. AAAI Conf. Artifical Intell.}, Honolulu, USA, Jul. 2019, pp. 8803--8810.

\bibitem{bib:TCNN}
K.~Kang \emph{et al.}, ``T-CNN: Tubelets with convolutional neural networks for object detection from videos'', \emph{IEEE Trans. Circuits Syst. Video Technol.}, vol. 28, no. 10, pp. 2896--2907.

%\bibitem{bib:FGFA}
%X.~Zhu, Y.~Wang, J.~Dai, L.~Yuan, and Y.~Wei, ``Flow-guided feature aggregation for video object detection,'' in \emph{Proc. Int. Conf. Comput. Vis.}, Venice, Italy, Oct. 2017, pp. 408--417.

\bibitem{bib:HPVD}
X.~Zhu, J.~Dai, L.~Yuan, and Y.~Wei, ``Towards high performance video object detection,'' in \emph{Proc. IEEE Conf. Comput. Vis. Pattern Recognition}, Salt Lack City, USA, Jun. 2018, pp. 7210--7218.

\bibitem{bib:STSN}
G.~Bertasius, L.~Torresani, and J.~Shi, ``Object detection in video with spatiotemporal sampling networks,'' in \emph{Proc. Eur. Conf. Comput. Vis.}, Munich, Germany, Sept. 2018, pp. 342--357.

%\bibitem{bib:STMN}
%F.~Xiao and Y.~J.~Lee, ``Video object detection with an aligned spatial-temporal memory,'' in \emph{Proc. Eur. Conf. Comput. Vis.}, Munich, Germany, Sept. 2018, pp. 494--510.

%\bibitem{bib:LSTM_SSD}
%M.~Liu and M.~Zhu, ``Mobile video object detection with temporally-aware feature APs,'' in \emph{Proc. IEEE Conf. Comput. Vis. Pattern Recognition}, Salt Lack City, USA, Jun. 2018, pp. 5686--5695.

\bibitem{bib:TSSD}
X.~Chen, J.~Yu, and Z.~Wu, ``Temporally identity-aware SSD with attentional LSTM,'' \emph{IEEE Trans. Cybern.}, doi:10.1109/TCYB.2019.2894261.

\bibitem{bib:TPN}
K.~Kang \emph{et al.}, ``Object detection in videos with tubelet proposal networks,'' in \emph{Proc. IEEE Conf. Comput. Vis. Pattern Recognition}, Hawaii, USA, Jul, 2017, pp. 727--735.

%\bibitem{bib:CloseLoop}
%L.~Galteri, L.~Seidenari, M.~Bertini, and A.~Del Bimbo, ``Spatio-temporal closed-loop object detection,'' \emph{IEEE Trans. Image Process.}, vol. 26, no. 3, pp. 1253--1263, 2017.

\bibitem{bib:TRN}
X.~Chen, J.~Yu, S.~Kong, Z.~Wu, and L.~Wen, ``Joint Anchor-Feature Refinement for Real-Time Accurate Object Detection in Images and Videos,'' \emph{arXiv:1807.08638}, 2018.

\bibitem{bib:CDT}
H.~U.~Kim and C.~S.~Kim, ``CDT: Cooperative detection and tracking for tracing multiple objects in video sequences. in \emph{Proc. Eur. Conf. Comput. Vis.}, Amsterdam, Netherlands, Oct. 2016, pp. 851--867.

%\bibitem{bib:Grab}
%A.~Saxena, J.~Driemeyer, and A.~Y.~Ng, ``Robotic grasping of novel objects using vision,'' \emph{Int. J. Robot. Res.}, vol. 27, no. 2, pp. 157--173, 2018.

\bibitem{bib:VOC}
M.~Everingham, L.~Van~Gool, C.~K.~Williams, J.~Winn, and A.~Zisserman, ``The pascal visual object classes (voc) challenge,'' \emph{Int. J. Comput. Vis.}, vol. 88, no. 2, pp. 303--338, 2010.

\bibitem{bib:SORT}
A.~Bewley, Z.~Ge, L.~Ott, F.~Ramos, and B.~Upcroft, ``Simple online and realtime tracking,'' in \emph{IEEE Int. Conf. Image Process.}, Phoneix, U.S., Sep. 2016, pp. 3464--3468.

%\bibitem{bib:MDP}
%Y.~Xiang, A.~Alahi, and S.~Savarese, ``Learning to track: Online multi-object tracking by decision making,'' in \emph{Proc. IEEE Int. Conf. Comput. Vis.}, Santiago, Chile, Dec. 2015, pp. 4705--4713.

\bibitem{bib:ALSTM}
Y.~Lu, C.~Lu, and C.~K.~Tang, ``Online video object detection using association LSTM,¡± in \emph{Proc. IEEE Int. Conf. Comput. Vis.}, Venice, Italy, Oct. 2017, pp. 2344--2352.

%\bibitem{bib:ROLO}
%G.~Ning, Z.~Zhang, C.~Huang, X.~Ren, H.~Wang, C.~Cai, and Z.~He, ``Spatially supervised recurrent convolutional neural networks for visual object tracking,'' in \emph{Proc. IEEE Int. Symp. Circuits Syst.}, Baltimore, USA, May 2017, pp. 1--4.

%\bibitem{bib:RelationMOT}
%J.~Xu, Y.~Cao, Z.~Zhang, H.~Hu, ``Spatial-temporal relation networks for multi-object tracking,'' \emph{arXiv:1904.11489}, 2019.

\bibitem{bib:ClearMOT}
K.~Bernardin and R.~Stiefelhagen, ``Evaluating multiple object tracking performance: The CLEAR MOT metrics,'' \emph{EURASIP J. Image Video Process.}, vol. 1, pp. 1--10, 2008.

\bibitem{bib:KCF}
J.~.F~Henriques, R.~Caseiro, P.~Martins, J.~Batista, ``High-speed tracking with kernelized correlation filters,'' \emph{IEEE Trans. Pattern Anal. Mach. Intell.}, vol. 37, no. 3, pp. 583-596, 2014.

%\bibitem{bib:SiamFC}
%L.~Bertinetto, J.~Valmadre, J.~F.~Henriques, A.~Vedaldi, and P.~H.~Torr, ``Fully-convolutional siamese networks for object tracking,'' in \emph{Proc. Eur. Conf. Comput. Vis.}, Amsterdam, Netherlands, Oct. 2016, pp. 850--865.

\bibitem{bib:SiamRPN}
B.~Li, J.~Yan, W.~Wu, Z.~Zhu, X.~Hu, ``High performance visual tracking with siamese region proposal network,'' in \emph{Proc. IEEE Conf. Comput. Vis. Pattern Recognition}, Salt Lack City, USA, Jun. 2018, pp. 8971--8980.

\bibitem{bib:SiamFC++}
Y.~Xu, Z.~Wang, Z.~Li, Y.~Yuan, and G.~Yu, ``SiamFC++: Towards robust and accurate visual tracking with target estimation guidelines,'' \emph{arXiv:1911.06188}, 2019.

%\bibitem{bib:SiamRPN++}
%B.~Li, W.~Wu, Q.~Wang, F.~Zhang, J.~Xing, J.~Yan, ``Siamrpn++: Evolution of siamese visual tracking with very deep networks,'' in \emph{Proc. IEEE Conf. Comput. Vis. Pattern Recognition}, Long Beach, USA, Jun. 2019, pp. 4282--4291.

%\bibitem{bib:DWSiamRPN}
%Z.~Zhang, H.~Peng, ``Deeper and wider siamese networks for real-time visual tracking,'' in \emph{Proc. IEEE Conf. Comput. Vis. Pattern Recognition}, Long Beach, USA, Jun. 2019, pp. 4591-4600.

%\bibitem{bib:SiamMask}
%Q.~Wang, L.~Zhang, L.~Bertinetto, W.~Hu, P.~H.~Torr, ``Fast online object tracking and segmentation: A unifying approach,'' in \emph{Proc. IEEE Conf. Comput. Vis. Pattern Recognition}, Long Beach, USA, Jun. 2019, pp. 1328--1338.

\bibitem{bib:TDR}
L.~Pang, Z.~Cao, J.~Yu, P.~Guan, X.~Rong, H.~Chai, ``A visual leader-following approach with a TDR framework for quadruped robots,'' \emph{IEEE Trans. on Syst., Man, and Cybern. Syst.}, doi: 10.1109/TSMC.2019.2912715, 2019.

\bibitem{bib:Stable}
H.~Zhang, N.~Wang, ``On the stability of video detection and tracking,'' \emph{arXiv:1611.06467}, 2016.

%\bibitem{bib:FasterRCNN}
%S.~Ren, K.~He, R.~Girshick, and J.~Sun, ``Faster R-CNN: Towards real-time object detection with region proposal networks,'' in \emph{Proc. Adv. Neural Info. Process Syst.}, Montreal, Canada, Dec. 2015, pp. 91--99.

%\bibitem{bib:RFCN}
%J.~Dai, Y.~Li, K.~He, and J.~Sun, ``R-FCN: Object detection via region-based fully convolutional networks,'' in \emph{Proc. Adv. Neural Info. Process Syst.}, Barcelona, Spain, Dec. 2016, pp. 379--387.

%\bibitem{bib:ACoupleNet}
%Y.~Zhu \emph{et al.}, ``Attention couplenet: Fully convolutional attention coupling network for object detection,'' \emph{IEEE Trans. Image Process.,} vol. 28, no. 1, pp. 113-126, 2019.

% YOLO
%\bibitem{bib:YOLO}
%J.~Redmon, S.~Divvala, R.~Girshick, and A.~Farhadi, ``You only look once: Unified, real-time object detection,'' in \emph{Proc. IEEE Conf. Comput. Vis. Pattern Recognition}, Las Vegas, USA, Jun. 2016, pp. 779--788.

% SSD
\bibitem{bib:SSD}
W.~Liu, D.~Anguelov, D.~Erhan, C.~Szegedy, S.~Reed, C.~Y.~Fu, and A.~C.~Berg, ``SSD: Single shot multibox detector,'' in \emph{Proc. Eur. Conf. Comput. Vis.}, Amsterdam, Netherlands, Oct. 2016, pp. 21--37.

\bibitem{bib:RetinaNet}
T.~Y.~Lin, P.~Goyal, R.~Girshick, K.~He, and P.~Dollar, ``Focal loss for dense object detection,'' in \emph{Proc. IEEE Int. Conf. Comput. Vis.}, Venice, Italy, Oct. 2017, pp. 2980--2988.

\bibitem{bib:RefineDet}
S.~Zhang, L.~Wen, X.~Bian, Z.~Lei, S.~Z.~Li, ``Single-shot refinement neural network for object detection,'' in \emph{Proc. IEEE Conf. Comput. Vis. Pattern Recognition}, Salt Lack City, USA, Jun. 2018, pp. 4203--4212.

\bibitem{bib:DRN}
X.~Chen, X.~Yang, S.~Kong Z.~Wu, and J.~Yu, ``Dual refinement network for single-shot object detection,'' in \emph{Proc. Int. Conf Robot. Autom.}, Montreal, Canada, May 2019, pp. 8305--8310.

\bibitem{bib:ATOM}
M.~Danelljan, G.~Bhat, F.~S.~Khan, M.~Felsberg, ``Atom: Accurate tracking by overlap maximization,'' in \emph{Proc. IEEE Conf. Comput. Vis. Pattern Recognition}, Long Beach, USA, Jun. 2019, pp. 4660--4669.

\bibitem{bib:FewShot}
A.~Li, T.~Luo, Z.~Lu, T.~Xiang, and L.~Wang, ``Large-scale few-shot learning: Knowledge transfer with class hierarchy,'' in \emph{Proc. IEEE Conf. Comput. Vis. Pattern Recognition}, Long Beach, USA, Jun. 2019, pp. 7212-7220.


\end{thebibliography}
\end{document}